\documentclass[preprint,12pt]{elsarticle}

\usepackage[top=1in,bottom=1in,left=0.75in,right=0.75in]{geometry}

\usepackage{longtable}
\usepackage{graphicx}
\usepackage{amssymb}
\usepackage{amsmath}
\usepackage{float}   

\journal{ }

\begin{document}

\begin{frontmatter}

\title{Limits of $n$-gram Style Control for LLMs via Logit-Space Injection}

\author{
  Sami-ul Ahmed\\
  University of Colorado Boulder\\
  \texttt{ahmed.samiul.h@gmail.com}
}
\begin{abstract}
Large language models (LLMs) are typically personalized via prompt engineering or parameter-efficient fine-tuning such as LoRA. However, writing style can be difficult to distill into a single prompt, and LoRA fine-tuning requires computationally intensive training and infrastructure. We investigate a possible lightweight alternative: steering a frozen LLM with $n$-gram style priors injected in logit space at decoding time. We train an $n$-gram model on stylistically distinct corpora -- including Don Quixote, CNN/DailyMail news headlines, and arXiv abstracts -- constructing an interpolated 1-to-3-gram prior over next-token probabilities. During generation we modify the LLM's logits by adding a weighted sum of style log-probabilities from each $n$-gram order that matches the current context, scaled by a control parameter $\lambda \in [0, 1]$.

We sweep $\lambda$ and style corpora and report style perplexity under the $n$-gram model, base-model perplexity as a proxy for fluency, Jensen–Shannon (JS) divergence between the original and steered token distributions, and token-overlap statistics. On TinyLlama-1.1B we identify a single narrow regime (for the \emph{Don Quixote} corpus at $\lambda=0.1$) where style perplexity improves by $24.7\%$ and base-model perplexity improves by $51.4\%$ relative to the frozen model. Outside this regime, and for multi-author corpora such as CNN/DailyMail and arXiv abstracts, even small nonzero $\lambda$ values generally result in worse style and fluency, and larger $\lambda$ values lead to collapse with extreme perplexities and incoherent text. Logit-space injection of $n$-gram style priors provides lightweight, tunable style control, but it is fragile: it operates effectively only within a narrow range of low $\lambda$ values and is consistently outperformed by prompting and LoRA. Prompting alone yields significant stylistic alignment while also substantially improving base-model perplexity, whereas LoRA is the most reliable and effective method overall. \end{abstract}


\end{frontmatter}

\section{Introduction}
\label{S:intro}

LLMs are widely used for writing tasks where we care not just about correctness but also about style, such as matching a user’s tone, vocabulary, and syntactic habits. Today, this is usually handled by fine-tuning models on specific data or writing prompts that describe the target style. Fine-tuning introduces extra parameters, infrastructure, and privacy concerns while prompting can be brittle and hard to control quantitatively. Therefore, we pose the question: Is there a lightweight, at-decoding method to steer style without training the base model? 

$n$-gram models have been used in related work, learning token-by-token from a corpus to predict the next token from the $n-1$ tokens before it. We define a style prior: an $n$-gram model trained using the target language model's tokenizer on a body of text. The style prior is used to tell us: based on these past $n-1$ tokens, what is most likely to come next under the target corpus.

In this work we explore a third option: using a style prior to steer LLM outputs in logit space: the multidimensional vector space containing the raw prediction scores for the next token before they are converted into probabilities. We train a small $n$-gram model on a corpus with a distinct style and use it to bias the LLM’s logits before softmax. This leaves the LLM frozen and offers a single scalar knob $\lambda$ to increase the strength with which the logits are being modified. This method requires no computationally expensive training as it leaves the base model untouched.

Similar methods have been explored in the past. Messner \& Lippincott (2025)~\citep{messner-lippincott-2025-transferring} use $n$-gram models to select dialect-specific subword realizations, which they used to generate text in the style of AAVE. Plug-and-Play Language Models (PPLM) apply gradients from an attribute classifier to the hidden state of a frozen LM to control sentiment or topic at inference time~\citep{dathathri2020plug}. Other work such as DeRa and PAD integrate reward-model scores during decoding to control alignment~\citep{liu2024dera, chen2024pad}. In many of these cases, control is achieved through neural critics trained on supervised preference data. In contrast, we investigate a purely non-parametric steering signal based on $n$-gram statistics extracted from a style corpus, and we study the effects of increasing the strength of the style prior on text generations.

Selected corpora include \emph{Don Quixote}, CNN/DailyMail headlines, and arXiv abstracts. Each of these offers a distinct style for us to attempt to emulate. If a corpus is too large to reasonably compute $n$-grams on, we train our model on a randomly sampled portion of it. We train our $n$-gram style prior using 1-, 2-, and 3-grams. 

We sweep across $\lambda$ values and style corpora and evaluate performance with style perplexity under an $n$-gram LM, base-model perplexity as a proxy for fluency, JS divergence, and token-overlap metrics. We compare these generations with prompt-only generation and LoRA fine-tuning.

We find that simple prompting works surprisingly well, sharply improving style alignment and fluency. LoRA aligns incredibly well to style while substantially improving fluency. Logit-space steering is tunable but fragile. Small $\lambda$ values yield minor style improvements on certain corpora but increasing $\lambda$ does not improve style alignment. Increasing $\lambda$ above a certain threshold (around $\lambda\approx0.6$) leads to collapse of model fluency; larger values of $\lambda$ fail to improve the model's adherence to the target style. Using sampling during inference reveals brittleness in the method, as some unlikely or incoherent tokens become boosted. At high $\lambda$ values, generations under both sampling and greedy decoding often become incoherent or fall into repetitive loops.

Our contributions are:

\begin{itemize}
  \item We formulate an $n$-gram style prior over the LLM tokenization and inject it directly into the model’s logits via a sparse, context-dependent logit update.
  \item We systematically evaluate style-control methods using perplexity-based metrics, JS divergence across decoding steps, and token-overlap statistics.   
  \item On TinyLlama-1.1B, we show that there is a very narrow range of $\lambda$ where style and fluency can simultaneously improve for one of three corpora, and that larger $\lambda$ values systematically destabilize the model, leading to extreme perplexities and incoherent text, contradicting the intuition that stronger priors always yield stronger style.
    \item The most effective style control mechanism remains LoRA fine-tuning, with prompting still providing significant control, and logit-space style injection offering only modest improvements within a narrow range of $\lambda$. 

\end{itemize}

Our results indicate that logit-space steering is highly sensitive to corpus complexity and only yields improvements within a low $\lambda$ regime on single-author data.

\section{Methods}
\label{S:methods}
\subsection{Style Prior}
\label{SS:style-prior}

The style prior is constructed through a sliding-window Markov-style probability construction. The model will "learn" how likely a token is after its previous $n-1$ tokens, essentially assembling a large table of conditional probabilities. For each order $n\in\{1, 2, 3\}$:
\[
n\text{-gram counts: } C_n(\text{context, token})
\]
\[\text{Context counts: } N_n(\text{context})=\sum_\text{token}C_n(\text{context, token})
\]
We estimate smoothed conditional probabilities:
\[
P_n(\text{token} \mid \text{context}) = \dfrac{C_n(\text{context, token})+k}{N_n (\text{context}) + kV}
\]
where $k$ is a small smoothing constant and $V$ is the vocabulary size. We set $k=10^{-3}$ and truncate each context's table to the top $K=512$ tokens to retain the most probable outcomes. This allows us to estimate the probability of a token occurring after a particular context.

We also assemble $P_{mix}$:

\[
P_{mix}(\text{token} \mid \text{context}) = \dfrac{\sum_n w_nP_n(\text{token} \mid \text{context)}}{\sum_n w_n}
\]
We use mixture weights $w_1= 0.1$, $w_2=0.3$, and $w_3=0.6$ to prioritize higher-order $n$-grams as they capture more specific stylistic and syntactic patterns than lower order ones. This weighting scheme assumes that tokens appearing after longer sequences (e.g. trigrams) provide a stronger signal of the style compared to shorter, more generic sequences. 
When computing $P_{mix}$, only $n$-gram orders for which the given context exists are included in the sum; we fall back to a small uniform distribution when no estimate exists. We use $\log P_{mix}$ when computing style perplexity.   
\subsection{Injection Mechanism}
\label{SS:injection-mechanism}

We modify the LLM's logits during decoding with the following mechanism. Let 
\\

\indent $z_i$: LLM logits for token $i$ pre-softmax\\
\indent$\lambda$: Parameter to control the strength of steering\\
\indent$w_n$: The weight for the $n$-gram of size $n$\\
\indent$P_n(i \mid \text{context})$: The $n$-gram probability that token $i$ comes after the context

\[
z_i' = z_i + \lambda \sum_n w_n\log P_n(i \mid \text{context})
\]

The update is sparse\footnote{While the definition of $P_{mix}$ in section 2.1 is a linear probability mixture, the logit injection is a log-linear combination}: at each decoding step we only modify logits for tokens that appear in the $n$-gram tables for the current context, leaving other logits unchanged. This method treats each $n$-gram order as an independent log-likelihood contributor and is computationally efficient since we are just performing arithmetic on logits before softmax, and the LLM stays frozen. We test various $\lambda$ values to gauge the effect of the style prior.

$\lambda$ values were intentionally selected within the range $[0, 1]$. $\lambda=0$ corresponds to where no style is injected, while $\lambda=1$ gives a more strongly style-biased distribution. Values $\lambda < 0$ would actively repel the model from the style prior, which is outside the scope of style alignment. Values $\lambda>1$ can allow the prior to dominate the logits and empirically destabilize decoding without improving alignment. Therefore we limit the sweep to $\lambda \in [0, 1]$, a stable and interpretable range. 

\section{Metrics}
We evaluate generations using style perplexity, base-model perplexity, JS divergence, and token-overlap metrics. 

\subsection{Style Perplexity}
\label{SS:style-perplexity}
We compute perplexity of the generated text under the interpolated style prior:
\[
\text{Style PPL}
= \exp\!\left(
-\frac{1}{T}
\sum_{t=1}^{T}
\log P_{mix}(x_t \mid x_{<t})
\right).
\]
Lower style perplexity indicates greater style alignment. 
\subsection{Base-Model Perplexity}
\label{SS:base-model-perplexity}
We also compute perplexity of the generated text under the base language model, allowing us to see if the generation is something that the language model would actually produce, which in turn allows us to see if the text is actually fluent and coherent. 
\subsection{Distribution Divergence}
\label{SS:distribution-divergence}
We compute Jensen-Shannon (JS) divergence as a metric for the distribution between base logits and biased logits across $\lambda$. JS divergence is computed across each decoding step and averaged across all tokens. This allows us to quantify the change in logits due to the style prior. 
\subsection{Token-Overlap Metrics}
\label{SS:token-overlap}
To evaluate surface-level stylistic vocab/phrase alignment, we measure unigram overlap rate and bigram seen rate. 
\begin{enumerate}
    \item Unigram Overlap Rate: fraction of generated tokens that appear in the top $K=5000$ unigrams from the style corpus. 
    \item Bigram Seen Rate: fraction of generated bigrams that are observed at least once in the style corpus. 
\end{enumerate}

\section{Experimental Setup}
\label{S:experimental-setup}

\subsection{Models and Inference}
\label{SS:models-inference}

The LLM chosen in this experiment was TinyLlama-1.1B, as its compact size allows rapid iteration with limited resources. Tokenization was done using the model's WordPiece tokenizer. Inference was performed on Google Colab using Python with the T4 GPU runtime, with FP16 and gradients disabled.

For each style corpus we construct the style prior on the model's tokenization. For Don Quixote we use the English translation and train on the full text. For CNN/DailyMail we use the headline portion of the dataset and subsample $200,000$ headlines for efficiency. For arXiv we sample $200,000$ paper abstracts from the `gfissore/arxiv-abstracts-2021` dataset. The prompts used for evaluation are generic and are not drawn from any of these corpora. 

We evaluate the effects of the style prior using greedy decoding, with generations spanning the values:
\[
\lambda \in \{0.00, 0.05, 0.10, 0.15, 0.20, 0.25, 0.30, 0.40, 0.50, 0.60, 0.70, 0.80, 0.90, 1.00\}
\] 
across all prompts, with the output set to a fixed length of 256 tokens. 20 prompts were constructed from a breadth of categories: \\
\subsection*{Narrative Prompts}
\begin{enumerate}
    \item ``I still remember the moment when everything began to change''
    \item ``At the edge of the city, far from the noise and lights''
    \item ``She had promised herself she would never return to this place''
    \item ``By the time anyone noticed the mistake, it was already too late''
    \item ``The evening air carried a quiet sense of anticipation''
\end{enumerate}

\subsection*{Dialogue Prompts}
\begin{enumerate}
    \item ``Are you sure this is the right decision?'' he asked
    \item ``If we don’t act now, we may lose our only chance,'' she replied
    \item ``That’s not what I meant,'' they said patiently
    \item ``Listen carefully, because I won’t repeat this again,''
    \item ``Look, here's the thing nobody wants to admit,''
\end{enumerate}

\subsection*{Expository Prompts}
\begin{enumerate}
    \item There are three main reasons why this issue is important.
    \item At first glance, it might seem that nothing unusual is happening.
    \item In recent years, many people have argued that this trend is accelerating.
    \item From a practical point of view, the situation can be summarized as follows.
    \item However, this explanation leaves out an important detail:
\end{enumerate}

\subsection*{Technical Prompts}
\begin{enumerate}
    \item First, we outline the basic idea of the method.
    \item The system consists of three main components:
    \item The goal of this section is to show that the proposed approach is effective.
    \item If we compare these two approaches, we find that several key differences emerge.
    \item To understand this more clearly, consider the following example:
\end{enumerate}

These 20 prompts span narrative, dialogue, expository, and technical styles to avoid overfitting to a single genre and to probe how the style prior behaves across qualitatively different inputs.

Unless otherwise noted, we use greedy decoding to isolate the effect of the style prior from sampling randomness. To evaluate whether the results were a cause of greedy decoding we repeat selected experiments on the arXiv dataset using top-p sampling ($p=0.9$, temperature $1.0$) and compare trends.

\subsection{Baseline Comparisons}
\label{SS:baseline-comparisons}

We benchmark the style injection by comparing to text generations of prompt-only style steering and a LoRA fine-tune upon the corpus.

\begin{itemize}
    \item Prompt-only Style Steering: We prompt the LLM with high level instructions to mimic the style of the corpus. 
    \begin{itemize}
        \item This was done for \emph{Don Quixote}
        \item Prompt: "Write in the style of Miguel de Cervantes in \emph{Don Quixote}"
    \end{itemize}
    \item LoRA fine-tune: We train a low-rank adaptation (LoRA) fine-tune with rank 8, $\alpha=16$, and dropout 0.05 on CNN/DailyMail headlines. We fine-tune TinyLlama-1.1B for 800 steps with batch size $8$ and learning rate $2\times10^{-4}$ in fp16, using only the headline text as supervision. This provides a strong style-specialized baseline for the news corpus.
\end{itemize}
This allows us to compare our method with existing methods of stylistic alignment. 

\section{Results}
\label{S:Results}

We present a quantitative analysis of logit-space steering across three distinct corpora. We use Jensen-Shannon (JS) Divergence to track the magnitude the distribution shifts by due to the style prior. We measure alignment of the style prior using Style Perplexity. We use Base-Model Perplexity to assess the impact of the style prior on linguistic fluency. Our results reveal that low values of $\lambda$ can improve stylistic fit in specific contexts, but higher injection strengths lead to a rapid degradation of model output coherence, eventually resulting in a complete collapse of fluency.

To quantitatively analyze our findings, we begin with an analysis of JS divergence across all decoding steps to see if the style prior is affecting the generations, and how changing $\lambda$ affects it. 

\subsection{Steering Behavior}
\label{SS:js-divergence}
\begin{figure}[H]
    \centering
    \includegraphics[width=0.75\linewidth]{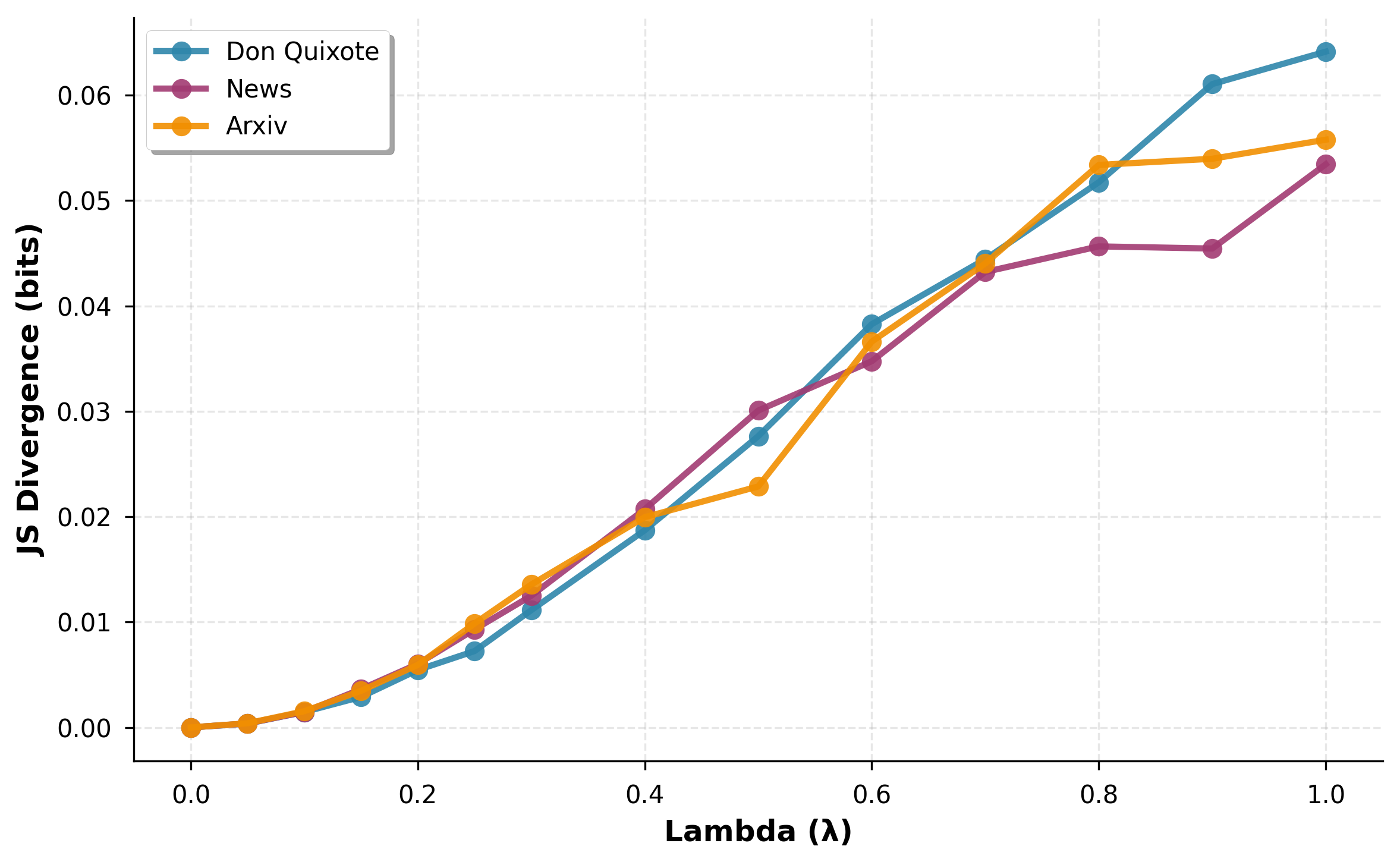}
    \caption{Steering strength increases with $\lambda$ (average across all decoding steps)}
\end{figure}

JS divergence is near 0 at $\lambda=0$ and grows roughly monotonically with $\lambda$ for all three corpora, reaching at most $\approx 0.06$ bits at $\lambda=1.0$ (0.0641 for \emph{Don Quixote}, 0.0534 for news, and 0.0558 for arXiv) (Figure 1). This shows that $\lambda$ provides a continuous, well-behaved control parameter in distribution space: larger $\lambda$ values reliably induce larger deviations between the base and modified distributions.

\subsection{Fluency Cost}
\label{SS:base-lm-ppl}
\begin{figure}[H]
    \centering
    \includegraphics[width=0.75\linewidth]{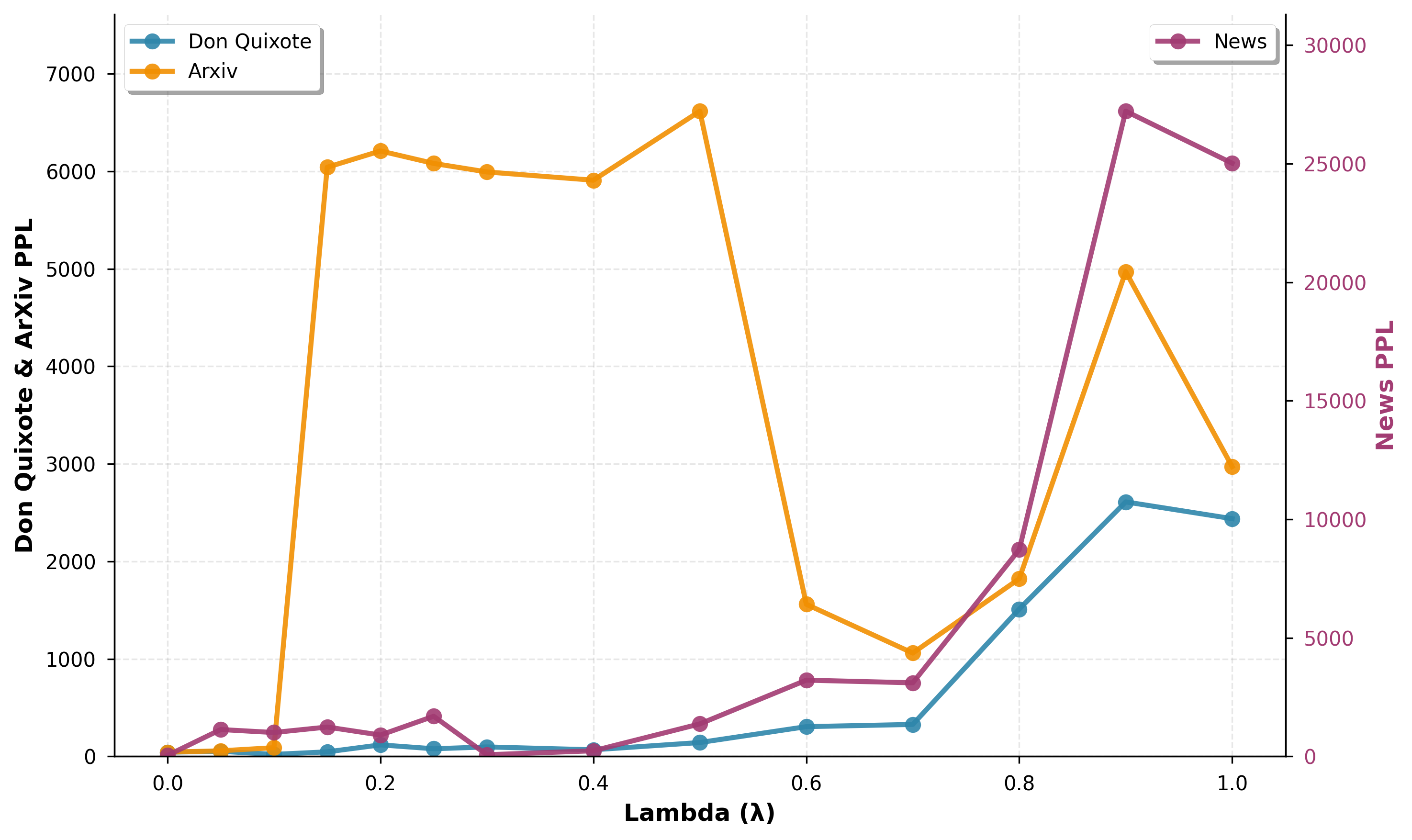}
    \caption{Fluency collapses at high $\lambda$}
\end{figure}
Fluency is robust only near $\lambda=0$ (and $\lambda=0.1$ for \emph{Don Quixote}). For \emph{Don Quixote}, base-model perplexity improves from $43.4$ at $\lambda=0$ to $21.1$ at $\lambda=0.1$, but then degrades, exceeding $1,500$ at $\lambda=0.8$ and $2,400$–$2,600$ at $\lambda \in \{0.9,1.0\}$. For the news corpus, many small nonzero $\lambda$ values already yield base PPL in the $900$–$1,700$ range, and high $\lambda$ values reach $25,000$–$27,000$. The arXiv corpus is even more fragile, with $\lambda \ge 0.15$ producing base PPL above $6,000$ in our greedy runs (Figure 2).

The relationship between $\lambda$ and base-model perplexity is nonlinear and heavily context dependent. For arXiv, the style prior almost immediately creates a spike in base-model perplexity, indicating an immediate loss of fluency. For \emph{Don Quixote} and news headlines corpora, the base-model perplexity follows a different trend: it remains stable up until $\lambda \approx 0.6$, and that the collapse regime begins at $\lambda>0.6$. All corpora indicate exponential increases in base-model perplexity, confirming that high-strength steering eventually overpowers the base model's coherence.

\subsection{Style Fit}
\label{SS:style-fit}
\begin{figure}[H]
    \centering
    \includegraphics[width=0.75\linewidth]{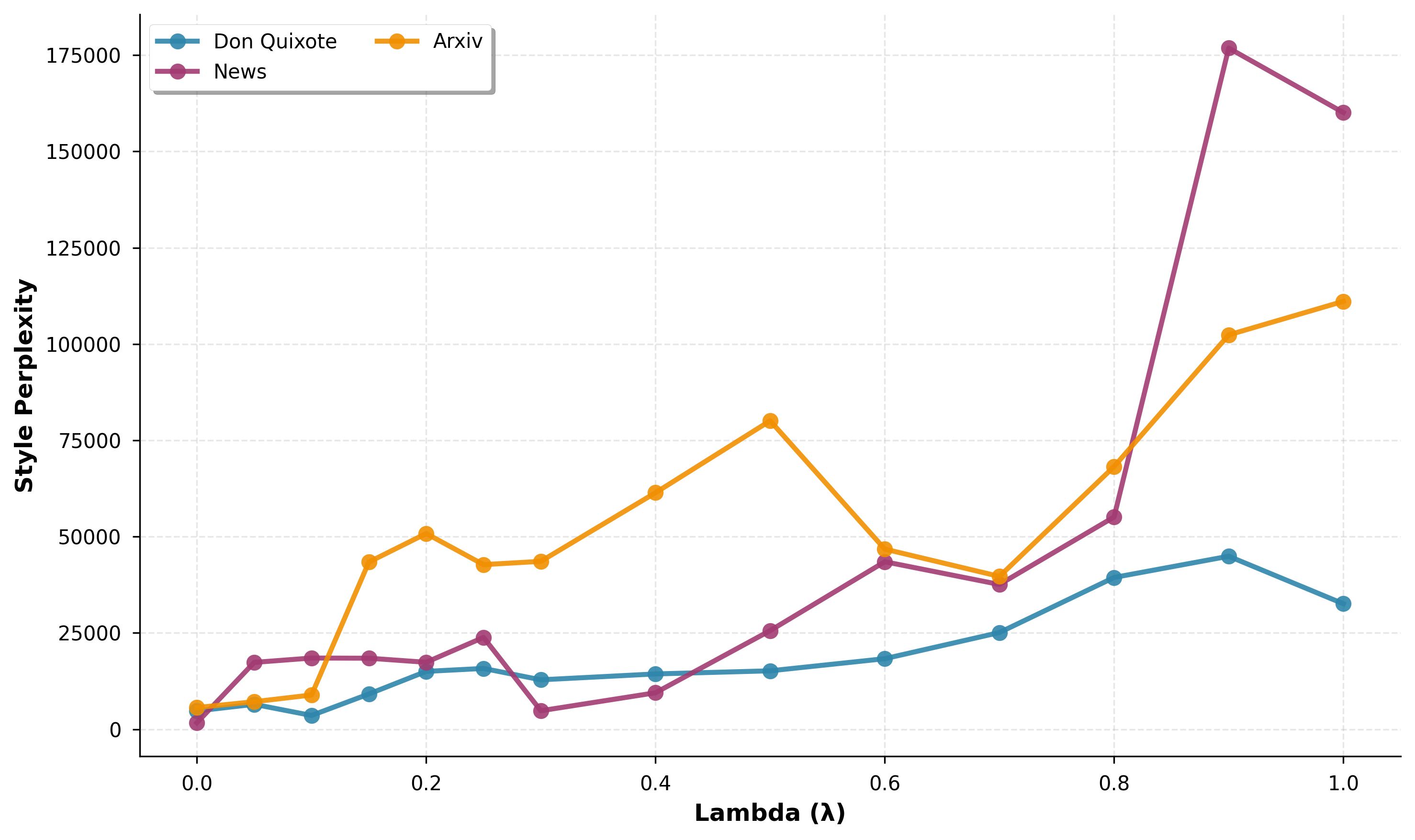}
    \caption{Higher $\lambda$ does not lead to "more style"}
\end{figure} 
Quantitative evaluation of stylistic alignment reveals that logit-space injection is incredibly fragile. Across all corpora and $\lambda$ values, we identify a single setting where the style prior successfully improves style perplexity: for \emph{Don Quixote} at $\lambda=0.1$, style perplexity decreases from $4751.3$ at $\lambda=0$ to $3577.7$, a $24.7\%$ improvement (Figure 3). 

Outside of this regime, increasing the steering strength ($\lambda$) is counterproductive:
\begin{itemize}
    \item Non-Monotonicity: Style perplexity does not decrease as $\lambda$ increases; instead, it sharply trends upward for all corpora. 
    \item Magnitude of Failure: For the news and arXiv corpora, any non-zero $\lambda$ immediately degrades stylistic fit. No value of $\lambda$ improves style perplexity for these corpora. Perplexity increases by factors of 10 to 100 relative to the baseline.
\end{itemize}

\subsection{Fluency-Style Tradeoff (Pareto Frontier)}
\label{SS:pareto-frontier}
\begin{figure}[H]
    \centering
    \includegraphics[width=1\linewidth]{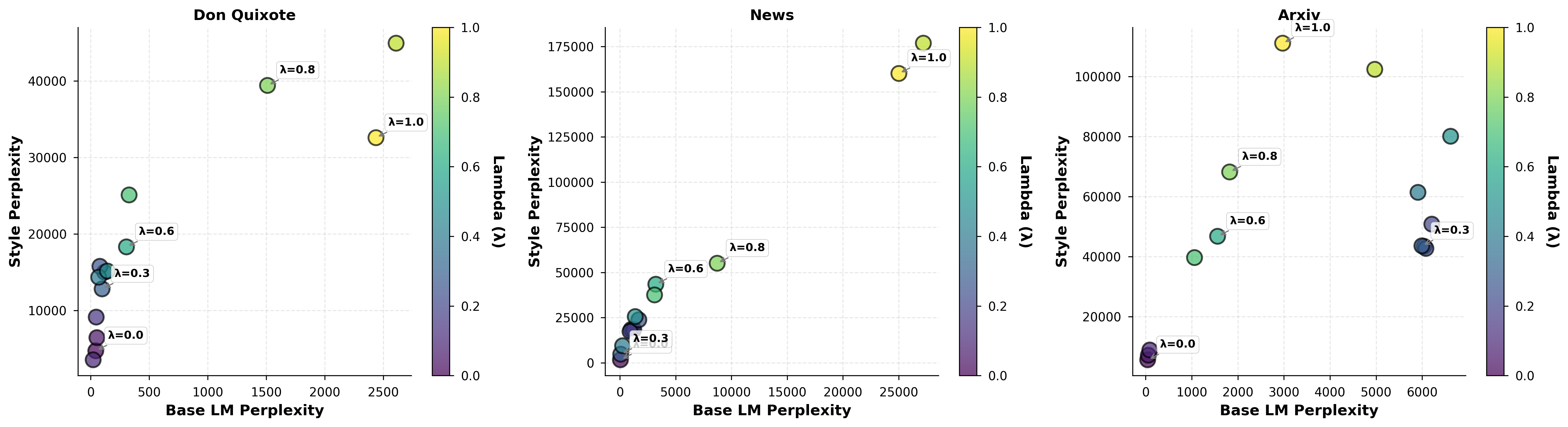}
    \caption{Higher $\lambda$ leads to collapse}
\end{figure}

The Pareto frontier in base-model perplexity–style perplexity space makes these trade-offs explicit (Figure 4). For \emph{Don Quixote}, the point corresponding to $\lambda=0.1$ ($21.1$ base-model perplexity, $3577.7$ style perplexity) strictly dominates the baseline $\lambda=0$ point ($43.4$, $4751.3$), forming a narrow area where both fluency and style improve. All larger $\lambda$ values move up and to the right, degrading one or both metrics. For the news and arXiv corpora, the $\lambda=0$ point ($43.4$, $1698.1$ for news; $43.4$, $5631.5$ for arXiv) is Pareto-optimal: every nonzero $\lambda$ yields higher style perplexity, higher base-model perplexity, or both. This visualization allows us to quantify the fragility of logit-space steering. While a "sweet spot" exists for the Don Quixote corpus at $\lambda=0.1$, the immediate divergence of the other two corpora proves that these improvements do not generalize.
\subsection{Lexical Structure Degradation}
\label{SS:lexical-structure}
\begin{figure}[H]
    \centering
    \includegraphics[width=1\linewidth]{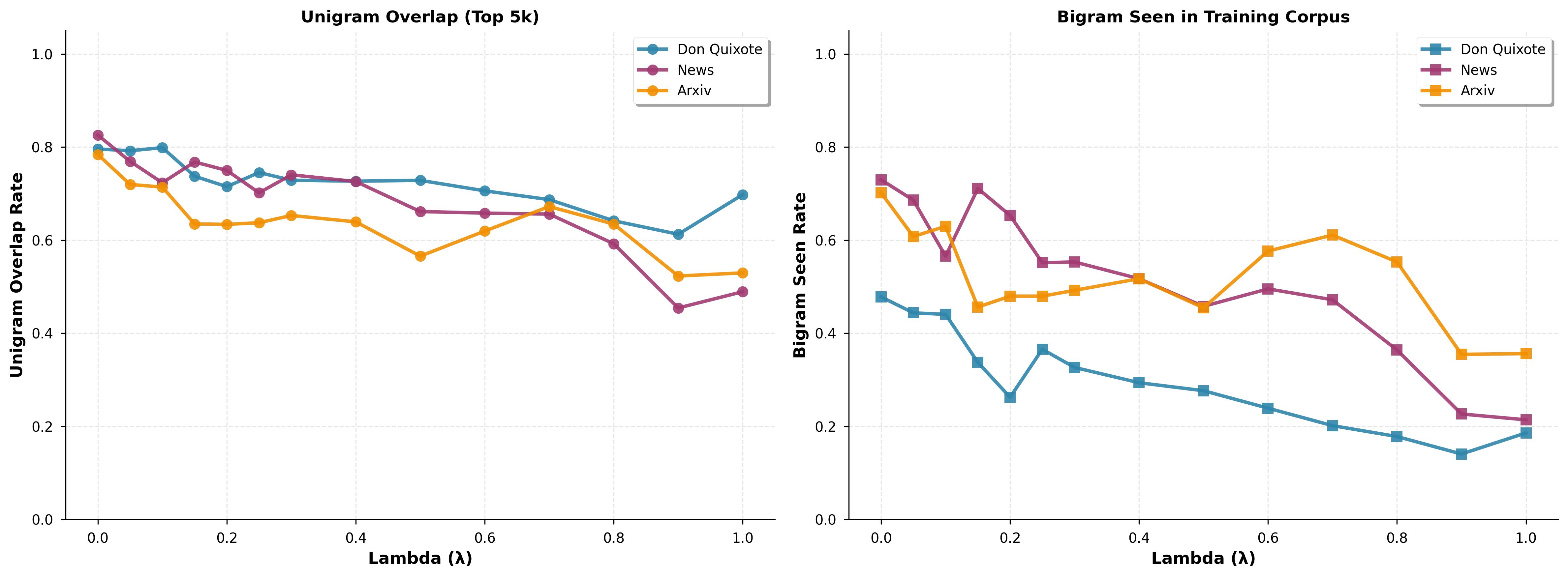}
    \caption{Lexical resemblance declines at high $\lambda$}
\end{figure}

While the previous spikes in perplexity indicate a loss of statistical alignment, the unigram overlap and bigram seen rate (Figure 5) explains the nature of the failure. As $\lambda$ increases, the model does not just struggle with style, it begins to diverge from the target vocabulary. 

For the news corpus, the unigram overlap rate decreases from $0.826$ at $\lambda = 0$ to $0.454$ at $\lambda=0.9$. This suggests that the model is generating tokens that do not exist in the desired style corpus. This general trend of decreasing unigram overlap rates is shared across the three corpora.

The sharp decline in bigram seen rate for the \emph{Don Quixote} corpus from $0.478$ to $0.141$ reveals that the model is not properly utilizing bigrams from the desired corpora. Part of this is due to the steering actually causing nonsensical repeating generations at high $\lambda$ (Table 1).

This reinforces previous results that stronger $n$-gram steering often yields generations that are less consistent with the desired style, and it adds the insight that part of the cause is unigram overlap and bigram seen rates decreasing.
\subsection{Baseline Comparison}
\label{SS:baseline}
\begin{figure}[H]
    \centering
    \includegraphics[width=1\linewidth]{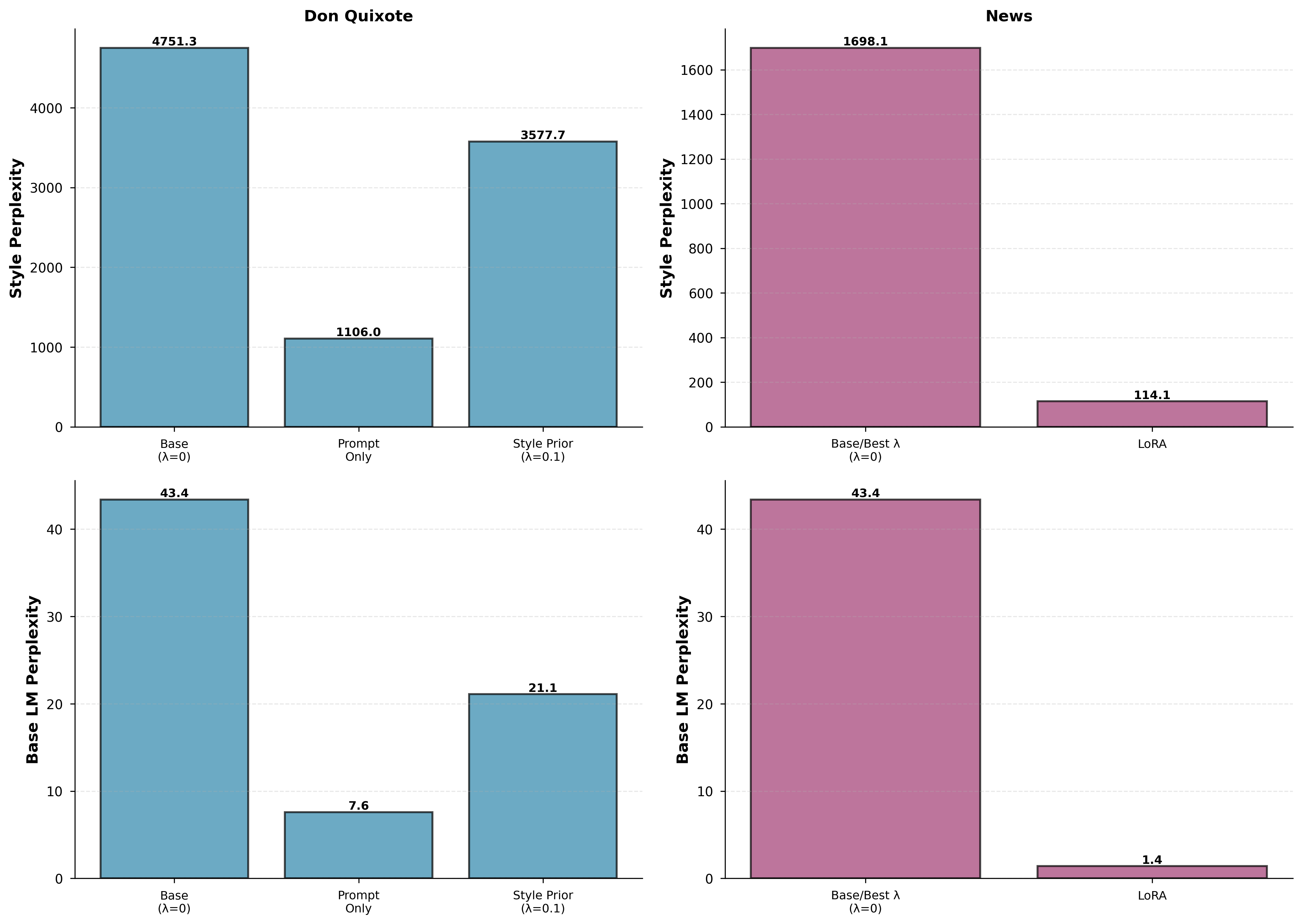}
    \caption{Comparing Decoding-Time Steering with Prompt-Only and LoRA}
\end{figure}

For \emph{Don Quixote}, the style perplexity is improved from baseline by $24.7\%$ with $\lambda=0.1$ (Figure 6). However, prompt-only modification demonstrates a much closer style fit to the target corpus and reduces base-model perplexity from $43.4$ to $7.58$. On the CNN/DailyMail news headlines and arXiv abstracts corpora, no $\lambda$ leads to a better style perplexity than $\lambda=0$. On the news corpus, LoRA yields an extremely strong style fit.

Prompt-only modification led to a $76.7\%$ improvement in style perplexity from baseline on \emph{Don Quixote} and an $82.5\%$ reduction in base-model perplexity, while LoRA led to a $93.3\%$ improvement in style perplexity and a $96.8\%$ reduction in base-model perplexity on the news corpus. Prompt-only modification is incredibly cheap yet still yields strong style alignment and fluency gains, but metrics-wise LoRA still outclasses it on the news domain.

The prompt-only and LoRA style methodologies lead to significantly better style alignment than any style prior tested.

\subsection{Sampling Robustness}
\label{SS:robustness}
The main experiments were performed using greedy decoding to rule out randomness as the cause of a generation and to isolate the effect of the style prior. We analyze one experiment over the arXiv dataset to see if similar trends follow when we use sampling. 
\begin{figure}[H]
    \centering
    \includegraphics[width=1\linewidth]{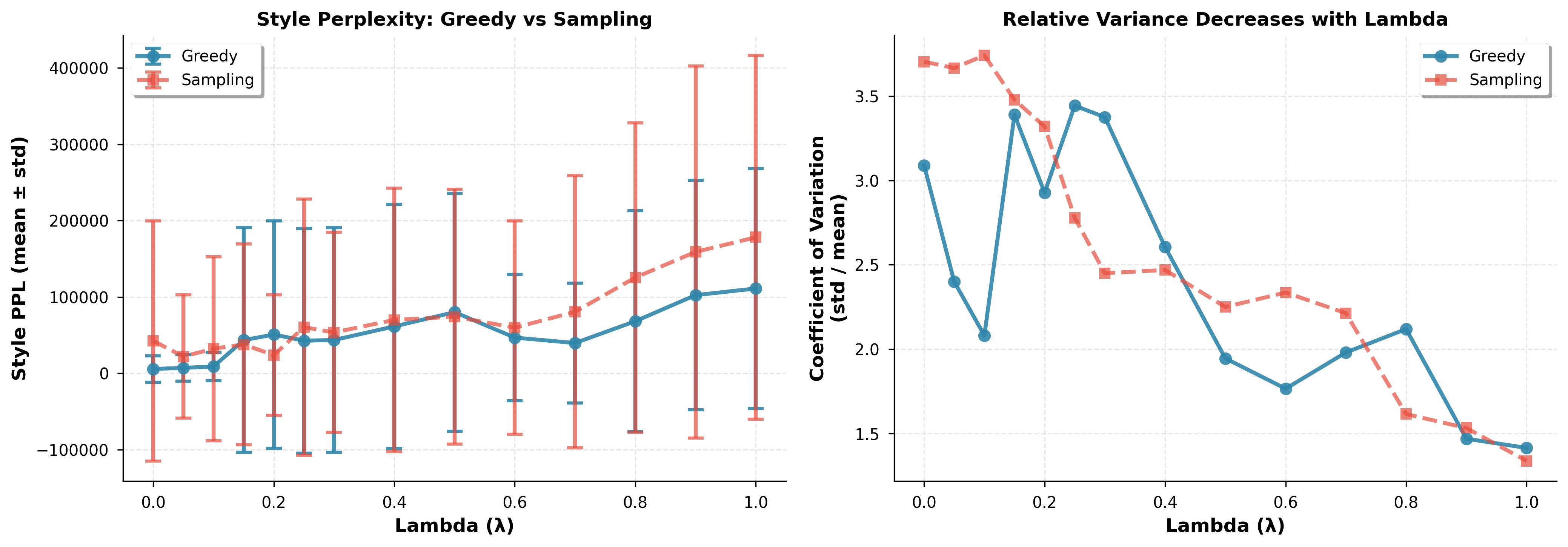}
    \caption{Sampling Robustness on arXiv Corpus}
\end{figure}

For arXiv, style perplexity under both greedy and sampling tends to increase with $\lambda$ (Figure 7). Under greedy decoding, mean style perplexity rises from $5631$ at $\lambda=0$ to over $1.1\times 10^5$ at $\lambda=1.0$, while the coefficient of variation (CV) drops from $3.09$ to $1.42$. Under sampling, the mean increases from $4.24\times10^4$ at $\lambda=0$ to $1.78\times10^5$ at $\lambda=1.0$, with CV decreasing from $3.70$ to $1.34$. This pattern suggests that stronger steering both worsens style perplexity and concentrates generations into fewer, more stereotyped modes. The qualitative trends in sampling with respect to $\lambda$ mirror those observed with greedy decoding.

\subsection{Qualitative Analysis}
\label{SS:qualitative}
We analyze a sample of generations across the \emph{Don Quixote} corpus.
{\small
\begin{longtable}{|l|r|r|p{0.6\textwidth}|}
\caption{Full Qualitative Examples Across Lambda Values}
\label{tab:qualitative-full}\\

\hline
\textbf{$\lambda$} & \textbf{Style PPL} & \textbf{Base PPL} & \textbf{Generated Text} \\
\hline
\endfirsthead

\hline
\textbf{$\lambda$} & \textbf{Style PPL} & \textbf{Base PPL} & \textbf{Generated Text} \\
\hline
\endhead

\hline
\multicolumn{4}{r}{Continued on next page}\\
\endfoot

\hline
\endlastfoot
\hline

0.0 & 784.1 & 1.9 & I still remember the moment when everything began to change. I was sitting in my room, staring at the ceiling, when I heard a knock on the door. I opened it to find my best friend, Sarah, standing there. "Hey, Emily," she said, smiling. "I heard you were feeling down. I wanted to come over and talk to you." I felt a wave of relief wash over me. Sarah was always there for me, no matter what. "Thanks, Sarah," I said, feeling a little nervous. "I'm just feeling a little down." Sarah sat down on the bed next to me and put her arm around me. "Emily, I know it's hard to talk about, but you're not alone. I've been through a lot, too. I'm here for you." I felt a lump form in my throat. Sarah had always been there for me, no matter what. "I'm sorry if I've been a little distant lately," I said, trying to keep my voice steady. "I've been busy with school and work, and I just needed some time to myself." \\
\hline

0.2 & 646.0 & 2.1 & I still remember the moment when everything began to change. I was sitting in my room, staring at the ceiling, lost in thought. I had just finished my final exam and was feeling a sense of relief wash over me. But as I looked up, I saw something that made my heart skip a beat. There, in the corner of the room, was a small, intricately designed flower. It was a delicate pink rose, with delicate petals that seemed to glow in the dim light. I couldn't believe my eyes. I picked up the rose and held it in my hand, marveling at its beauty. It was unlike anything I had ever seen before. The petals were soft and velvety, with a delicate fragrance that filled the room. As I looked closer, I realized that the rose was not just any ordinary flower. It was a symbol of hope, a reminder that even in the darkest of times, there is always something beautiful to be found. From that moment on, I began to see the world in a new light. I started to appreciate the small things, the moments that seemed insignificant at the time but that would later on become cherished mem \\
\hline

0.4 & 392.6 & 1.4 & I still remember the moment when everything began to change. I remember the feeling of the rain on my skin, the taste of the salt on my lips, and the warmth of the sun on my skin. I remember the sound of the waves crashing against the shore, the feel of the sand between my toes, and the smell of the ocean. I remember the feeling of the waves crashing against the shore, the feel of the sand between my toes, and the smell of the ocean. I remember the feeling of the rain on my skin, the taste of the salt on my lips, and the warmth of the sun on my skin. I remember the feeling of the rain on my skin, the taste of the salt on my lips, and the warmth of the sun on my skin. I remember the feeling of the rain on my skin, the taste of the salt on my lips, and the warmth of the sun on my skin. I remember the feeling of the rain on my skin, the taste of the salt on my lips, and the warmth of the sun on my skin. I remember the feeling of the rain on my skin, the taste of the salt on my lips, and the warmth \\
\hline

0.6 & 2632.4 & 1.7 & I still remember the moment when everything began to change. Scene 2: INT. KITCHEN - DAY Jenna is cooking dinner for her family, and Sarah walks in. JENNA Hi Sarah, dinner's ready. SARAH Thanks, Jenna. JENNA You're welcome. JENNA You're welcome. JENNA You're welcome. JENNA You're welcome. JENNA You're welcome. JENNA You're welcome. JENNA You're welcome. JENNA You're welcome. JENNA You're welcome. JENNA You're welcome. JENNA You're welcome. JENNA You're welcome. JENNA You're welcome. JENNA You're welcome. JENNA You're welcome. JENNA You're welcome \\
\hline

0.8 & 5077.8 & 2.1 & I still remember the moment when everything began to change. Scene 2: INT. KITCHEN - DAY Jane is cooking dinner for herself and her family, surrounded by the smell of garlic and tomato sauce. JANE (to herself) You're doing this, Jane. You're doing this. Jane's phone rings. JANE Hello, Jane's phone rings. Jane answers. JANE Hello, Jane's voice. Jane's husband, JOHN, answers. JOHN Hi Jane, Jane's husband, JOHN, answers. JOHN Hi Jane, Jane's husband, JOHN, speaks. JOHN We're sorry to bother you, Jane, Jane's husband, JOHN, speaks. JOHN We're sorry to bother you, Jane, Jane's husband, JOHN, speaks. JOHN We're sorry to \\
\hline

1.0 & 6826.5 & 2.9 & I still remember the moment when everything began to change. Scene 2: INT. KITCHEN - DAY Jane is cooking dinner for herself and her family, surrounded by the smell of garlic and tomato sauce. JANE "Mom, dinner's ready!" Jane calls out to her mother as she pours herself some water. MARILYN "Thank god, Jane!" Marilyn looks relieved. Jane's mother takes a sip of the warm water, and Jane watches as she starts to choke. MARILYN "Oh no!" Jane's mother gasps. Jane's mother tries to breathe but fails, and Jane's mother falls onto Jane's lap. JANE "Mom, what's happening?" Jane asks as tears stream down her face. MARILYN "Oh Jane, I'm so sorry. I've been drinking too much." Marilyn looks ashamed. Jane's mother looks defeated. JANE "Mom, I'm sorry too." \\
\hline

Prompt & 172.4 & 2.1 & Write in the style of Miguel de Cervantes in \emph{Don Quixote}. I still remember the moment when everything began to change. It was a warm summer evening, and I was sitting on the porch of my grandparents' house, reading a book. I was lost in the story, lost in the world of the characters, lost in the words. Suddenly, I heard a noise. It was a rustling in the bushes, and I knew that it was not a bird or a squirrel. It was a man. I stood up, my heart racing, and walked towards the sound. When I got closer, I saw that it was a man in a white cloak, with a long beard and a sword in his hand. He was walking towards me, and I could see that he was not alone. There were two other men with him, and they were also dressed in white. I was terrified. I had never seen anyone like this before. I had never seen a man with a sword before. I had never seen a man with a white cloak before. I had never seen a man with two other men with him before. The man with the sword stopped in front of me, and I could see that he was a knight. He was tall, with a long, flowing beard \\
\hline

\end{longtable}
}
We can see that in some high $\lambda$ generations, nonsensical repetition cycles occur (Table 1), likely due to the style prior elevating repeating tokens that usually would not be selected by the LLM.

\section{Discussion}
\label{S:discussion}

Our findings highlight that the style prior logit-space injection does not beat baseline methods of modifying style. Tweaking $\lambda$ produces a narrow sweet spot where metrics improve on \emph{Don Quixote}, but even so, prompt-only steering performs better, and LoRA remains substantially stronger overall (Figure 6). 

We see that small $\lambda$ outperforms large $\lambda$ in multiple metrics, likely because at higher $\lambda$ the probabilistic model attempts to boost tokens that make sense at the trigram level, but not within the general context. At lower $\lambda$ the LLM is able to "overpower" the style prior and keep generations on the right track (Figure 3). 

The $n$-gram prior and the LLM distribution become mathematically incompatible at high $\lambda$. The $n$-gram model is context-poor, while the LLM is context-rich. This forces the model to pick words that satisfy local statistical patterns, violating the LLM's global logic. Consequently, the LLM collapses into a $3$-gram repetition loop that it finds as the highest probability "style" move -- even as it destroys the global coherence of the generation (Table 1).    

The style prior performed the best on \emph{Don Quixote}. \emph{Don Quixote} has one author, whereas the CNN/DailyMail and arXiv abstracts datasets are aggregations of multiple authors and carry their distinct tones, making it difficult for the Markovian model to pick up. The arXiv abstracts and news headline datasets have high entropy because they contain technical jargon or diverse reporting styles which are hard to grasp through an $n$-gram model. \emph{Don Quixote} has a single, distinguishable style, allowing the model to learn stable patterns (Figure 3). The book contains archaic language patterns such as "thou art" which are very distinct from the base LLM's training data. Such patterns are easy to build an $n$-gram over and steer toward, even if fragile. 

Within our sweep, we did not find an optimal configuration that uses $n$-gram to capture the style of arXiv abstracts and news headlines, the Pareto-optimal points for both are at baseline ($\lambda$=0). This suggests that logit-space injection is not just worse, but results in degradation of both key metrics simultaneously (Figure 4). 

Additionally, unigrams, bigrams, and trigrams capture archaic repeated phrase structure, but they do not capture longer academic/news rhetorical structure. 

While our results demonstrate the fragility of logit-space injection, this exploration is essential for exploring the usability and boundaries of modular AI control. These findings help to identify possible areas where simple statistical priors can be valuable, while also contextualizing the method in the space of style control.

\section{Related Work}
\label{S:related}

Decoding-time control for LLMs has been approached from several angles. Plug-and-Play Language Models steer generation by applying gradients from an attribute classifier to the hidden state of a frozen LM, enabling sentiment or topic control at inference time without parameter updates~\citep{dathathri2020plug}. Critic-guided decoding uses a critic model trained in an actor-critic framework to reweight token logits during decoding according to predicted reward~\citep{kim2023critic}. Both methods demonstrate the fundamental trade-off between control strength and fluency.

More recent work has focused on using external signals to modify logits directly during decoding. DeRa introduces a continuous alignment knob by incorporating a reward-model alignment score into decoding~\citep{liu2024dera}. PAD extends this idea to personalization, training a user-specific reward model and applying it at inference~\citep{chen2024pad}. These approaches require neural critics trained on supervised preference data; by contrast, our method requires no additional learned components.

Several lines of research revisit classical $n$-gram models in the LLM era. Li et al.\ show that learning a neural language model on the residual between an $n$-gram model and the true distribution improves perplexity and rare-token prediction, indicating that $n$-grams retain valuable local structure~\citep{li2022ngramisback}. Infini-gram further demonstrates that massive, web-scale $n$-gram statistics remain effective and efficient to query~\citep{liu2024infinigram}. These results support our use of $n$-gram statistics as a stylistic prior rather than a full generative model.
Messner \& Lippincott (2025)~\citep{messner-lippincott-2025-transferring} also operate in logit space using $n$-gram statistics, but with a different goal: they construct an $n$-gram model from a target style corpus and use its probabilities to rescale token logits to make the LLM select dialect-specific subword realizations (e.g., variations in AAVE). Their method is designed explicitly for style and dialect transfer via extreme subword variation; their results demonstrate that $n$-gram-based logit-space steering can produce lightweight stylistic alignment without changing model weights. 

By contrast, we treat the $n$-gram model as a general-purpose Markov prior over the full next-token distribution and study its behavior as a tunable probabilistic component with $\lambda$. We do not optimize for dialect rewriting, and instead we focus on how increasing the strength of the style prior trades off style and fluency. Overall, we analyze the dynamics of $n$-gram style tuning. 

Our approach departs from prior work in three ways: (1) it is fully non-parametric, requiring no additional models or training; (2) it provides a continuous style-control knob analogous to DeRa’s alignment parameter but grounded in empirical token transitions; and (3) it reframes $n$-gram models as a plug-in personalization signal rather than a fallback LM or safety filter.
\section{Conclusion}
\label{S:conclusion}

We investigated stylistic control via logit-space injection of an $n$-gram prior. Our results demonstrate that this approach is fragile, consistently resulting in collapse at higher strengths. Logit-space injection is dominated by both prompt-only steering and LoRA fine-tuning across all tested regimes. In fact, logit-space injection worsens style and base-LM perplexity in all cases beyond a narrow stability regime: on \emph{Don Quixote} with $\lambda=0.1$ where style perplexity improved by 24.7\% (Figure 3).

The core finding of this experiment is that logit-space steering results in collapse of the generation. When applied to high-entropy corpora such as news headlines and arXiv abstracts, the baseline model with $\lambda=0$ remains the only Pareto-optimal configuration (Figure 4). Increasing steering at all for these corpora results in worse metrics, and at higher $\lambda$ values the model is forced into repetitive loops of nonsense where the generated text becomes highly implausible under both the target style and natural language (Table 1). Ultimately, while logit-space steering is computationally lightweight, it is Pareto-dominated by weight-altering methods due to its failure to capture longer-term structural patterns while maintaining global coherence. 

Future work can explore:

\begin{itemize}
    \item Better calibrated priors or an $n$-gram that looks further back.
    \item Incorporating neural models that can retain longer-term sequences to guide style, as done by Kim et al. (2023)~\citep{kim2023critic}
    \item Penalizing repeating and illogical generations.
    \item Further analysis of style prior performance on single-author corpora. 
\end{itemize}

\normalsize
\bibliographystyle{elsarticle-num}
\bibliography{references}

@inproceedings{dathathri2020plug,
  title     = {Plug and Play Language Models: A Simple Approach to Controlled Text Generation},
  author    = {Dathathri, Sumanth and Madotto, Andrea and Lan, Janice and Hung, Jane and Frank, Eric and Molino, Piero and Yosinski, Jason and Liu, Rosanne},
  booktitle = {International Conference on Learning Representations},
  year      = {2020},
  url       = {https://openreview.net/forum?id=H1edEyBKDS},
  note      = {arXiv:1912.02164}
}

@inproceedings{kim2023critic,
  title     = {Critic-Guided Decoding for Controlled Text Generation},
  author    = {Kim, Minbeom and Lee, Hwanhee and Yoo, Kang Min and Park, Joonsuk and Lee, Hwaran and Jung, Kyomin},
  booktitle = {Findings of the Association for Computational Linguistics: ACL 2023},
  pages     = {4598--4612},
  year      = {2023},
  address   = {Toronto, Canada},
  publisher = {Association for Computational Linguistics},
  doi       = {10.18653/v1/2023.findings-acl.281},
  url       = {https://aclanthology.org/2023.findings-acl.281/}
}

@inproceedings{messner-lippincott-2025-transferring,
  title     = {Transferring Extreme Subword Style Using Ngram Model-Based Logit Scaling},
  author    = {Messner, Craig and Lippincott, Tom},
  booktitle = {Proceedings of the 5th International Conference on Natural Language Processing for Digital Humanities},
  month     = may,
  year      = {2025},
  address   = {Albuquerque, USA},
  publisher = {Association for Computational Linguistics},
  pages     = {272--280},
  url       = {https://aclanthology.org/2025.nlp4dh-1.24/},
  doi       = {10.18653/v1/2025.nlp4dh-1.24}
}

@inproceedings{li2022ngramisback,
  title     = {$N$-gram Is Back: Residual Learning of Neural Text Generation with $n$-gram Language Model},
  author    = {Li, Huayang and Cai, Deng and Xu, Jin and Watanabe, Taro},
  booktitle = {Findings of the Association for Computational Linguistics: EMNLP 2022},
  year      = {2022},
  address   = {Abu Dhabi, United Arab Emirates},
  publisher = {Association for Computational Linguistics},
  pages     = {1523--1533},
  url       = {https://aclanthology.org/2022.findings-emnlp.109},
  doi       = {10.18653/v1/2022.findings-emnlp.109}
}

@article{liu2024infinigram,
  title   = {Infini-gram: Scaling Unbounded $n$-gram Language Models to a Trillion Tokens},
  author  = {Liu, Jiacheng and Min, Sewon and Zettlemoyer, Luke and Choi, Yejin and Hajishirzi, Hannaneh},
  journal = {arXiv preprint arXiv:2401.17377},
  year    = {2024},
  url     = {https://arxiv.org/abs/2401.17377}
}

@inproceedings{liu2024dera,
  title     = {Decoding-time Realignment of Language Models},
  author    = {Liu, Tianlin and Guo, Shangmin and Bianco, Leonardo and Calandriello, Daniele and Berthet, Quentin and Llinares, Felipe and Hoffmann, Jessica and Dixon, Lucas and Valko, Michal and Blondel, Mathieu},
  booktitle = {Proceedings of the 2024 Conference on Empirical Methods in Natural Language Processing},
  year      = {2024},
  publisher = {Association for Computational Linguistics},
  url       = {https://arxiv.org/abs/2402.02992}
}

@article{chen2024pad,
  title   = {PAD: Personalized Alignment of LLMs at Decoding-Time},
  author  = {Chen, Ruizhe and Zhang, Xiaotian and Luo, Meng and Chai, Wenhao and Liu, Zuozhu},
  journal = {arXiv preprint arXiv:2410.04070},
  year    = {2024},
  url     = {https://arxiv.org/abs/2410.04070}
}

\end{document}